\documentclass[times,twocolumn,final]{elsarticle}

\usepackage{cag}
\usepackage{framed,multirow}

\usepackage{amssymb}
\usepackage{latexsym}
\usepackage{booktabs}

\usepackage{url}
\usepackage{xcolor}
\definecolor{newcolor}{rgb}{.8,.349,.1}

\usepackage{hyperref}

\usepackage[switch,pagewise]{lineno} 

\journal{Computers \& Graphics}

\begin{document}

\verso{L. Zhao, F. Han and X. Peng et al.}

\begin{frontmatter}

\title{Cartoonish sketch-based face editing in videos using identity deformation transfer}%


\author[1]{Long Zhao\corref{cor1}}
\ead{lz311@cs.rutgers.edu}
\cortext[cor1]{Corresponding author at: Rutgers University, 110 Frelinghuysen Road, Piscataway, New Jersey, 08854-8019, United States.}
\author[1]{Fangda Han}
\author[2]{Xi Peng}
\author[1]{Xun Zhang}
\author[1]{Mubbasir Kapadia}
\author[1]{Vladimir Pavlovic}
\author[1]{Dimitris N. Metaxas}

\address[1]{Department of Computer Science, Rutgers University, NJ, United States}
\address[2]{Department of Computer Science, Binghamton University, New York, United States}

\received{9 December 2018}
\accepted{13 January 2019}

\begin{abstract}
We address the problem of using hand-drawn sketches to create exaggerated deformations to faces in videos, such as enlarging the shape or modifying the position of eyes or mouth. This task is formulated as a 3D face model reconstruction and deformation problem. We first recover the facial identity and expressions from the video by fitting a face morphable model for each frame. At the same time, user's editing intention is recognized from input sketches as a set of facial modifications. Then a novel identity deformation algorithm is proposed to transfer these facial deformations from 2D space to the 3D facial identity directly while preserving the facial expressions. After an optional stage for further refining the 3D face model, these changes are propagated to the whole video with the modified identity. Both the user study and experimental results demonstrate that our sketching framework can help users effectively edit facial identities in videos, while high consistency and fidelity are ensured at the same time.
\end{abstract}


\end{frontmatter}


\section{Introduction}\label{sec:introduction}

Recent years have witnessed tremendous advances in the field of facial performance capture in videos, which serves as a vital foundation for other computer graphics applications~\cite{han2017deepsketch2face, thies2016face2face, yang2012facial}. Especially, impressive results have been achieved in state-of-the-art face editing frameworks, and they are widely used in creating funny facial effects for video games, movies and even mobile applications. In order to express user's editing intention, this kind of frameworks always involves complex inputs (e.g., other images or videos within the same domain~\cite{yang2012facial, yang2011expression}) or additional capture devices (e.g., RGB or RGB-D cameras~\cite{thies2016face2face, chen2013accurate, hsieh2015unconstrained}). However, it is quite inconvenient for artists or amateur editors to reach these resources in our daily life. Moreover, current state-of-the-art methods always aim to enable users to modify the facial expression of the actor in a video, since this kind of editing intention can be easily detected with fixed facial identities. While changing the identity, i.e., the original appearance of a face without the influence from the pose and facial expression, is quite difficult, since it is a form of modification which is hard to be computed straightforward from reference inputs or several parameters.

\begin{figure*}[!t]
\centering
\includegraphics[width=\linewidth]{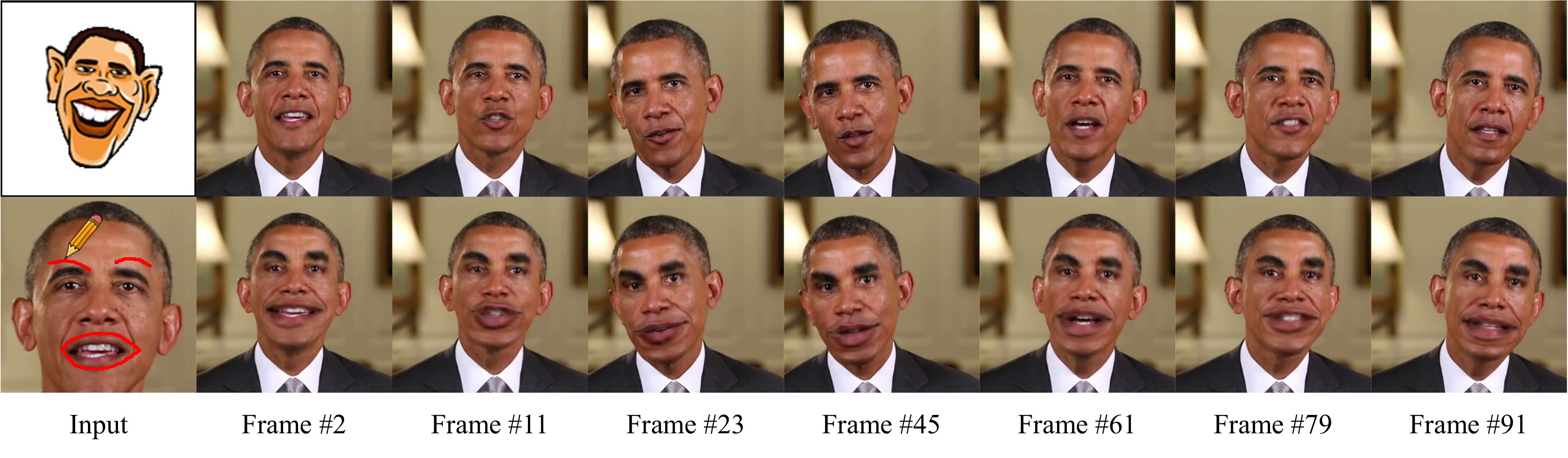}
\caption{Given a video, consider the problem of modifying the actor's facial appearance, e.g. enlarging his mouth, to match facial characters in a reference image (top-left). Our framework enables users to perform such editing operations by hand-drawn sketch. Then these modifications are propagated throughout the whole video. Top: a cartoon image for reference and key frames in a video. Bottom: the input sketches and edited frames. \emph{Note that our framework is not a caricature system. It does not transfer texture styles from cartoon images to videos, since we only focus on sketch-based identity deformation.}}
\label{fig:teaser}
\end{figure*}

We address these two shortcomings by making use of sketch, which offers more efficiency and flexibility to editors as demonstrated in the recent research~\cite{lau2009face, miranda2012sketch, wang2015sketch, liang2014sketch}. Consider the problem of transferring facial characters in a cartoon image to an actor in the video as shown in Fig.~\ref{fig:teaser}. In this paper, we propose a novel and robust interactive sketch-based face editing framework for both professional and amateur editors to finish this task very conveniently on a standard PC. \emph{We note that our framework is not a caricature system: the cartoon image we introduced here is not an input which will be processed by our framework; we are simply using it as an inspiration or guideline for users to edit the source video.} In fact, users are free to modify any facial appearance of an actor in the given video with the help of our framework. Compared to previous work, we focus on allowing users to edit the facial identity of the actor in the whole video, but not the expressions.

There are three challenges towards this goal. (1) There is an inherent tradeoff between flexibility of sketch-based specification and robustness. Specifically, unconstrained hand-drawn strokes may produce ambiguous inputs~\cite{lau2009face,chang2006sketching}. For example, a stroke drawn between the eyebrow and upper eyelid might indicate editing either of them. And it is quite difficult for the framework to determine the user's true editing intention by dealing with this stroke alone. (2) Since the face appearance depends on the pose of the actor as well as the identity, the influence of facial expression should be taken into account when applying changes to the identity. (3) Compared with previous sketch-based methods designed for static 2D images or 3D models~\cite{lau2009face,miranda2012sketch,chang2006sketching,nataneli2011bringing}, our framework has to further propagate the modifications from one frame to the whole video. In this process, we need to predict the modified face appearance in each frame while ensuring consistency and fidelity.

In this paper, we introduce a 3D face model fitting and identity deformation transfer formulation. Our core idea is to first transfer modifications from the input sketch to the corresponding 3D face model fitted by the facial identity and expression, which is then used for propagating changes to the whole video. To the best of our knowledge, we are the first framework which allows users to edit the facial identity of an actor in a video using hand-drawn sketches. This is made possible with the following key contributions: (1) a sketch-based facial identity editing framework for videos, (2) a novel 2D to 3D sketch-based identity deformation transfer algorithm, (3) and a contour-based interface for 3D model refinement.

\begin{figure*}[!t]
\centering
\includegraphics[width=\linewidth]{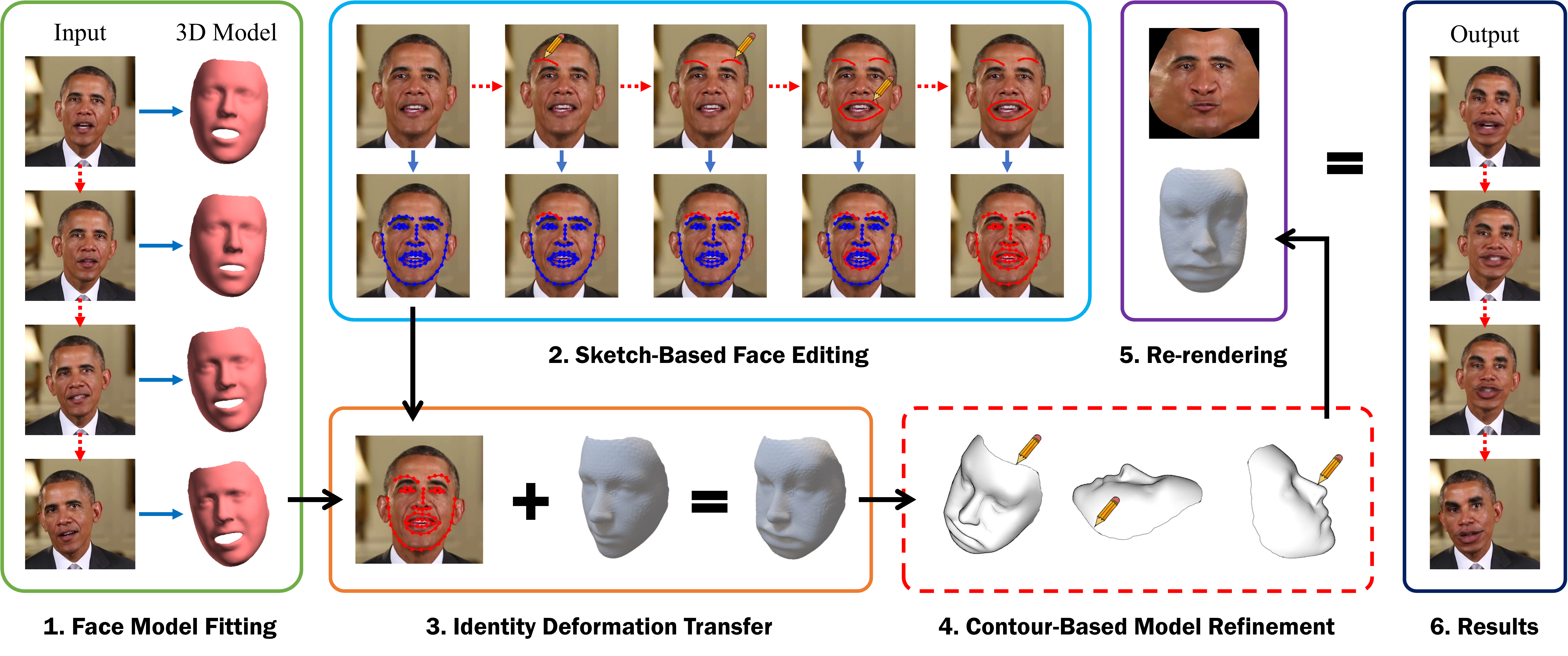}
\caption{Workflow of our sketching framework. The detailed algorithm of each part is presented in the corresponding sections. Red dashed arrows indicate a sequence of inputs while blue solid arrows stand for model fitting or updating operations, and a dashed box means an optional step.}
\label{fig:overview}
\end{figure*}

\section{Related work}\label{sec:relatedwork}

In this section, we review the previous work in the filed of sketch-based editing and deformation transfer which motivate the design of our sketching system.

\subsection{Sketch-based shape editing}

Hand-drawn sketches are widely used in modeling static facial inputs, such as images or 3D shapes~\cite{han2017deepsketch2face,olsen2009sketch}. The main challenge of these systems is to handle ambiguous user inputs, i.e., strokes which are difficult to match. Previous work~\cite{lau2009face,chang2006sketching,sucontphunt2008interactive}  limits the use of only pre-recorded data (curves or points) to mitigate ambiguity. The following two sketch-based facial animation editing systems proposed recently are most related to our work. Nataneli et al.~\cite{nataneli2011bringing} introduced an internal representation of sketch's semantics, while users have to draw sketch in some predefined regions. Miranda et al.~\cite{miranda2012sketch} built a sketching interface control system, which only allows users to draw strokes on a predefined set of areas corresponding to different face landmarks to avoid ambiguous conditions. In this paper, we introduce a sketch-based editing framework which differs from previous work in the following two aspects: (1) our method is the first framework that allows users to edit the face by sketch in a video sequence other than a static image or 3D shape; (2) we utilize the sequence information of strokes to deal with ambiguous user inputs without predefined constraints.

\subsection{Deformation transfer}

Deformation transfer~\cite{sumner2004deformation, botsch2006deformation} firstly addressed the problem of transferring local deformations between two different meshes, where the deformation gradient of meshes is directly transferred by solving an optimization problem. Semantic deformation transfer~\cite{baran2009semantic} inferred a correspondence between the shape spaces of the two characters from given example mesh pairs by using standard linear algebra. Zhou et al.~\cite{zhou2016cartoon} further utilized these methods to automatically generate a 3D cartoon of a real 3D face. Thies et al.~\cite{thies2016face2face} developed a system that transfers expression changes from the source to the target actor based on~\cite{sumner2004deformation} and achieves real-time performance. Xu et al.~\cite{xu2014controllable} designed a facial expression transfer and editing technique for high-fidelity facial performance data. Moreover, other flow-based approaches~\cite{yang2012facial, yang2011expression} are also proposed to transfer facial expression to different face meshes. However, these traditional methods aim to transfer deformations, especially facial expressions, between 3D meshes. Differing from them, we propose a transfer pipeline which can be used to directly transfer local identity changes in 2D space to a 3D face model. Huang et al.~\cite{huang2006subspace} presented an approach to project changes of a mesh in 2D to 3D as the projection constraint. Compared with it, the main novelty of our algorithm is that we combine a sketch-based interface to enable users to perform the editing with hand-drawn sketches from 2D to 3D. We first map sketch into a set of modifications corresponding to 3D space, and then transfer it to the target 3D mesh.

\section{Framework overview}\label{sec:overview}

\begin{table*}[t!]
\renewcommand{\arraystretch}{1.2}
\caption{\label{tbl:symbol}The main symbols and their meanings we employed in this paper.}
\centering
\begin{tabular}{@{}ll@{}}
\toprule
Symbol & Meaning \\ \midrule
$V = \{\mathbf{v}_k \mid 1 \leq k \leq N_V\}$ & a set of vertices to represent the shape of a 3D face mesh \\
$\mathbf{u} = [u_1, \dots, u_n]$ & the coefficient vector of the facial identity \\
$\mathbf{e} = [e_1, \dots, e_n]$ & the coefficient vector of the facial expression \\
$B = \{\mathbf{b}_k \mid 0 \leq k < N_B\}$ & a set of expression blendshapes of a certain actor ($\mathbf{b}_0$ is the shape of neutral expression)\\
$L = \{\mathbf{l}_k \mid 1 \leq k \leq N_L\}$ & a set of 2D landmarks of the face in a video frame \\
$F$ & a 3D face mesh generated after the global rotation $\mathbf{R}$ and translation $\mathbf{t}$ \\
$I$ & the identity component of a 3D face mesh $F$ \\
$E$ & the expression component of a 3D face mesh $F$ \\
$M$ & the isomap of a 3D face mesh $F$ \\
$\{G_k \mid 1 \leq k \leq N_G \}$ & a set of 2D landmark groups (each $G_k$ is a subset of $L$) \\
$\{S_k \mid 1 \leq k \leq N_S \}$ & a set of hand-drawn strokes \\
\bottomrule
\end{tabular}
\end{table*}

The input of our framework is a monocular video consisting of continuous frames of a person's face, together with a frame $t_0$ in this video containing a corresponding hand-drawn sketch. This sketch may be a complete facial sketch or partial strokes representing changes that the user wants to be made to the appearance of the face, e.g., to enlarge the mouth or modify the position of the eyebrows. Our goal is to recognize all these changes from the sketch and apply them to the whole video. Inspired by Thies et al.~\cite{thies2016face2face, yang2012facial}, we formulate this task as a parametric face model fitting and deformation transfer problem.

The whole pipeline of our framework is outlined in Fig.~\ref{fig:overview}. Our core idea is to first reconstruct a 3D face model $F_t$ for each frame $t$ in the input video, where $F_t$ can be disentangled into a unique component $I$ to represent the facial identity and a sequence of $E_t$ to describe the facial expression changes over time. Meanwhile, the face deformation encoded in the input sketch is approximated by a set of local deformations in 2D space. Then we transfer these deformations from 2D space to the target 3D facial identity $I$, while the influence of expression $E_t$ is removed by solving an energy minimization problem. After computing the modified identity shape $\hat{I}$, we can get the updated full 3D face model $\hat{F}_t$ for each frame $t$. Finally, these modifications are propagated throughout the whole video by rendering $\hat{F}_t$ to frame $t$ with the isomap $M_t$.

In the following, we discuss the individual steps in detail. First, in Section~\ref{sec:approach:fitting}, we show how the 3D face model is reconstructed and disentangled into the identity as well as expression for each frame in the video by a robust 3D face morphable model fitting algorithm~\cite{cao2014displaced}. In Section~\ref{sec:approach:sketch}, a robust sketch mapping and fitting schema is introduced to recognize user's editing intentions and apply them to the face in 2D space. Specifically, we utilize the order information carried by a series of strokes to mitigate ambiguity. Then an energy function is minimized to deform the face appearance and handle stroke noises at the same time. In Section~\ref{sec:approach:transfer}, we further present an approach to transfer deformations from 2D space to the 3D facial identity with depth estimation, while get rid of the influence from expression. We also implement a contour-based interface for users to refine the 3D identity optionally in Section~\ref{sec:approach:refinement}. Finally, Section~\ref{sec:approach:optimization} presents the optimization algorithms for rendering the deformed face texture back to the whole video, which removes artifacts generated from the previous steps. The user study as well as experiment results shown in Section~\ref{sec:results} indicate that our sketching framework is simple to use even for amateurs, while high-fidelity results are guaranteed as well. For clarity, we list the main symbols we used throughout this paper in Table~\ref{tbl:symbol}.

\section{Face model fitting}\label{sec:approach:fitting}

We utilize FaceWarehouse~\cite{cao2014facewarehouse} to construct the blendshape face meshes for each frame in the input video. Specifically, a fully transformed 3D face model $F$ can be represented as:
\begin{equation}
\label{eq:fullfacemodel}
F = \mathbf{R} \cdot V + \mathbf{t} = \mathbf{R} \cdot \big(C \times_2 \mathbf{u}^T \times_3 \mathbf{e}^T\big) + \mathbf{t},
\end{equation}
where $V$ is a set of vertices describing the shape of the face mesh. We utilize $\mathbf{R}$ and $\mathbf{t}$ to represent the global rotation and translation of $V$, respectively. $C$ is the rank-3 core tensor from the FaceWarehouse database~\cite{cao2014facewarehouse}, which corresponds to vertices of the face mesh, identity and expression. $\mathbf{u}$ is the identity vector and $\mathbf{e}$ is the expression vector. Let $L = \{\mathbf{l}_k\}$ denote a set of 2D facial landmarks of the face in a frame and $\{\mathbf{v}_k\}$ denote their corresponding vertices of the 3D face mesh $V$. Let $F^{(\mathbf{v}_k)}$ be the 3D position of $\mathbf{v}_k$ after the global rotation and translation according to Eq.~\ref{eq:fullfacemodel}. To compute $\mathbf{l}_k$, we define a set of 2D displacement $D = \{\mathbf{d}_k\}$ and each of them is added to the projection of $F^{(\mathbf{v}_k)}$:
\begin{equation}
\label{eq:projfacemodel}
\mathbf{l}_k = \Pi_\mathbf{P}\big(F^{(\mathbf{v}_k)}\big) + \mathbf{d}_k,
\end{equation}
where $\Pi_\mathbf{P}(\cdot)$ denotes a perspective projection operation which is parameterized by a projection matrix $\mathbf{P}$. To recover these unknown face parameters from the video, we employ DDE~\cite{cao2014displaced}, a state-of-the-art real-time regression algorithm for facial tracking. DDE predicts a sextuple $\langle \mathbf{P}, \mathbf{u}; \mathbf{e}_t, \mathbf{R}_t, \mathbf{t}_t, D_t \rangle$ for each frame $t$ in the video. Note that $\mathbf{P}$ and $\mathbf{u}$ are invariant across all frames for the same actor and the same video camera during tracking, while the other unknowns change in different frames. The expression blendshapes $B = \{\mathbf{b}_k\}$ of the certain actor in the input video is constructed as:
\begin{equation}
\label{eq:expfacemodel}
B = C \times_2 \mathbf{u}^T.
\end{equation}
As commonly assumed in blendshape models, $\mathbf{b}_0$ is 3D shape of the neutral face. We can further represent the face shape $V_t$ of the actor in the frame $t$ by:
\begin{equation}
\label{eq:shapefacemodel}
V_t = \mathbf{b}_0 + \sum_{n=1}^{N_B - 1}(\mathbf{b}_n - \mathbf{b}_0) \cdot \hat{e}_t^{(n)} = \mathbf{b}_0 + E_t = I + E_t,
\end{equation}
where $\hat{\mathbf{e}}_t = [\hat{e}_t^{(1)},\dots,\hat{e}_t^{(n)}]$ is computed from the initially fitted expression coefficient vector $\mathbf{e}_t$ estimated by DDE. Intuitively, Eq.~\ref{eq:shapefacemodel} disentangles the actor's 3D face shape $V_t$ into the identity $I$ and expression $E_t$. For each frame $t$, we also extract $M_t$, the isomap~\cite{tenenbaum2000global} which contains pixel textures for the face model, and compute 2D landmarks $L_t$ according to Eq.~\ref{eq:projfacemodel}. At last, we get the final per-frame face performances which consist of six parameters $\langle I; E_t, \mathbf{R}_t, \mathbf{t}_t, L_t, M_t \rangle$. In the following steps, $I$, $E_t$ and $L_t$ are used for computing the deformed 3D face mesh; while $\mathbf{R}_t$, $\mathbf{t}_t$ and $M_t$ are employed to propagate the modifications throughout the entire video to generate the final result.

\section{Sketch-based deformation transfer}\label{sec:approach:sketch}

We present a robust sketch-based face editing framework to enable users to edit all possible face details once they have been marked with the corresponding vertices on the 3D face mesh. In this paper, we allow users to edit 68 face landmarks predefined by Cao et al.~\cite{cao2014displaced,huber2016multiresolution} for illustration as shown in Fig.~\ref{fig:sketch}(b). In order to apply user's editing intention from the input sketch, we need first map each stroke to a suitable part of the face, e.g., the contour of eyes or mouth, and then deform this part according to the hand-drawn stroke.

\subsection{Sketch mapping}\label{sec:approach:sketch:mapping}

\begin{figure}[!t]
\centering
\includegraphics[width=\linewidth]{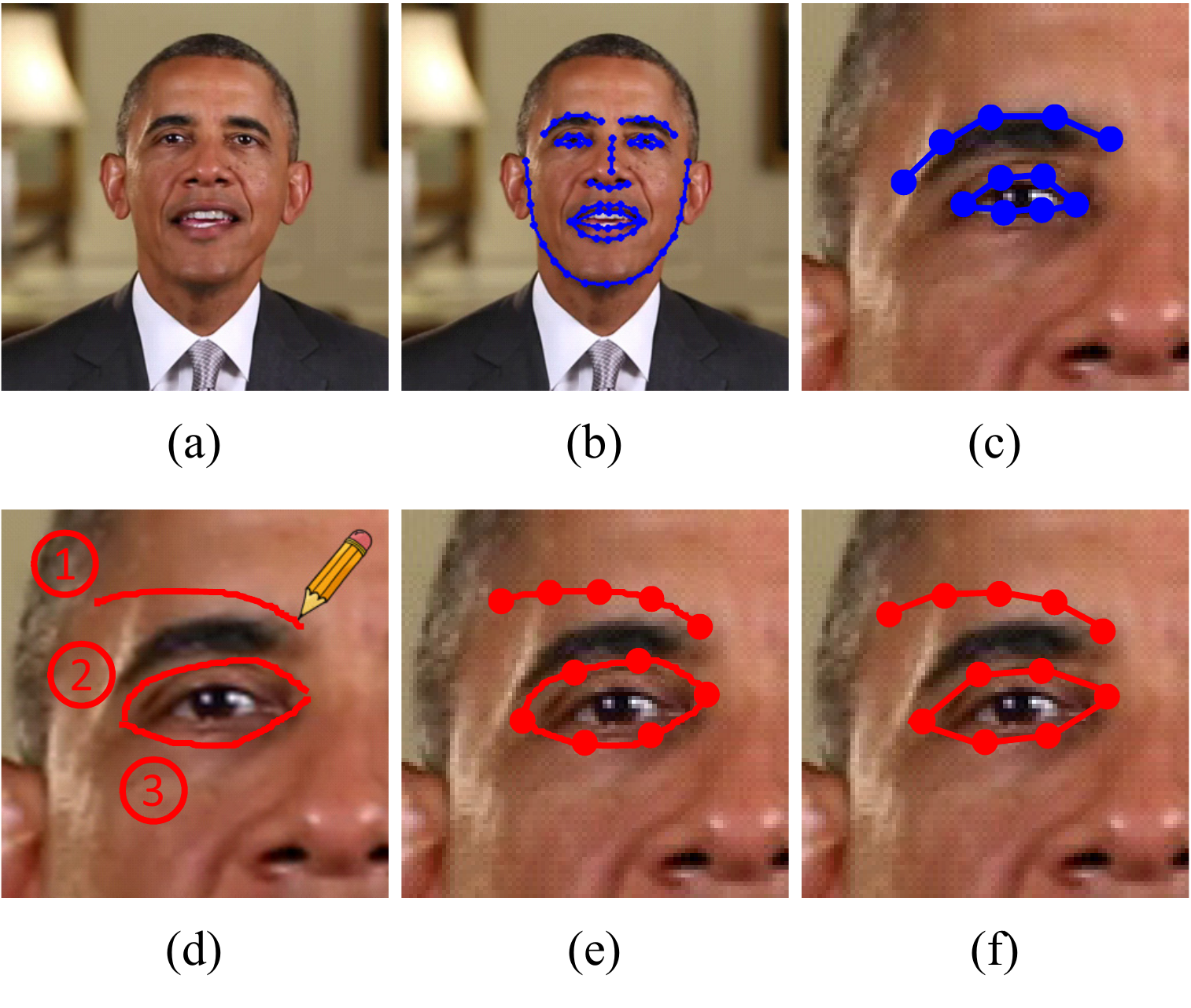}
\caption{Illustration of the face landmarks. (a) the original frame; (b) the 68 landmarks predefined by Huber et al.~\cite{huber2016multiresolution}; (c) the original landmarks to be edited; (d) the sketch input (the second stroke is ambiguous since it can be mapped to either the eyebrow or the upper eyelid); (e) key points extracted from the strokes; (f) the final modified landmarks.}
\label{fig:sketch}
\end{figure}

Our target is to map each stroke to a landmark group in the 2D space (a collection of landmarks which represents a meaningful part of the face, e.g., the left eyebrow or the upper eyelid), and to remove unreasonable strokes from the result at the same time. The main challenge in this task is how to deal with ambiguous user inputs, and Fig.~\ref{fig:sketch}(d) shows an example. Previous methods solve this problem by only allowing users to edit the face with pre-defined curves~\cite{lau2009face,chang2006sketching,sucontphunt2008interactive} or draw strokes in pre-ordered regions~\cite{miranda2012sketch, nataneli2011bringing}. Instead, we introduce a robust mapping schema which enables users to draw strokes without certain constraints, while the only assumption we made here is that the stroke should be ``clean'', i.e., each stroke aims to edit just one target landmark group.

We notice users always draw a sketch in a meaningful order encoding their editing intention. Landmark groups having a strong relation with each other, e.g., the upper and bottom eyelids of the same eye, trend to be drawn at the same time. Based on this observation, the input sketch is regarded as an ordered sequence of strokes and Hidden Markov Model~(HMM) is employed to formulate this problem. \emph{Note that our HMM-based algorithm leverages the order in which users draw strokes, for efficient matching with landmark groups. Prior works, e.g.,~\cite{kraevoy2009modeling} have not combined this order information with HMM.} It uses HMM to model the shape of this stroke to match mesh templates. Let $\{G_1, \dots, G_t\}$ be a set of landmark groups on the 2D image, and $\{S_1, \dots, S_t\}$ be the stroke sequence of the input sketch. We treat each landmark group $G$ as the hidden state while a stroke $S$ is the observation of HMM, and our target is to find the most probable sequence of hidden states (landmark groups) for a given observation sequence (strokes):
\begin{equation}
\label{eq:sketch:mapping:seq}
\mathop{\arg\max}_{G_{1:t}}P(G_{1:t}|S_{1:t}).
\end{equation}
The initial probabilities $P(G_0)$ for each hidden state is set to $1/N_G$. And $P(S_t|G_t)$ is the emission probability which measures the probability of each stroke belonging to a certain landmark group. We define $P(S_t|G_t)$ as:
\begin{equation}
P(S_t|G_t) =
  \left\{\begin{array}{l@{\qquad}l}
    \exp(-\frac{d(S_t, G_t)^2}{2\sigma^2}), & \textrm{if } d(S_t, G_t) \leq 3\sigma \\[\jot]
    0,              & \textrm{otherwise}
  \end{array}\right.
\label{eq:sketch:mapping:emission}
\end{equation}
where $d(S_t, G_t)$ measures the difference between $S_t$ and $G_t$, which is the average Euclidean distance of their corresponding key points. Note that if $S_t$ has a high distance with all landmark groups (which means that this stroke is invalid), $S_t$ will not be matched with any $G_t$ and excluded from the result. $P(G_t|G_{t - 1})$ is the transition matrix which expresses the probability of moving from one hidden state to another. Transition probabilities between two landmark groups with a strong relation are assigned a higher value, i.e., twice as large as the other values, which makes it easier for corresponding strokes to be mapped together and helps when strokes are ambiguous. Given these parameters, the most probable sequence problem can be solved by the Viterbi Algorithm~\cite{viterbi1967error}.

\subsection{Landmark deformation}\label{sec:approach:sketch:fitting}

For each input stroke $S$ and its mapped landmark group $G$, we need to deform $G$ into $\hat{G}$ according to $S$, where $\hat{G}$ is the final modified landmark group. This is achieved by solving an energy minimization function that leverages the position of the input stroke (editing intention of the user) and the original shape of the landmark group. Let $\mathbf{g}_i$ and $\hat{\mathbf{g}}_i$ be the coordinates of the $i$-th landmark in $G$ and $\hat{G}$, respectively, and $\mathbf{s}_i$ be the corresponding $i$-th key point in $S$. The energy function is formulated as:
 \begin{equation}
 \label{eq:sketch:fitting:target}
\mathcal{E} = \sum_{i = 1}^{N_G}||\hat{\mathbf{g}}_i - \mathbf{s}_i||^2 + \sum_{i = 1}^{N_G - 1}(1 - \cos(\hat{\gamma}_i - \gamma_i)),
\end{equation}
where $\gamma_i$ is the included angle of $\mathbf{g}_i$ and $\mathbf{g}_{i + 1}$; $\hat{\gamma}_i$ is that of $\hat{\mathbf{g}}_i$ and $\hat{\mathbf{g}}_{i + 1}$. To minimize this target function, we use the value of $||\mathbf{g}_{i + 1} - \mathbf{g}_{i}||$ to approximate $||\hat{\mathbf{g}}_{i + 1} - \hat{\mathbf{g}}_{i}||$, and we solve it with gradient descent algorithm.

Intuitively, the first term of Eq.~\ref{eq:sketch:fitting:target} is the position constrain. It measures the distance between the modified landmark group and the input stroke, which moves the landmarks of this landmark group to their expected positions. Meanwhile, the second term of Eq.~\ref{eq:sketch:fitting:target} represents the shape prior. It is employed to maintain the original shape information of this landmark group after the modification, which helps to prevent generating unrealistic results due to noises carried by the input sketch.

\section{Facial identity deformation transfer}\label{sec:approach:transfer}

The final modified facial identity $\hat{I}$ is calculated from the target identity $I$ by transferring 2D deformations (a set of modified 2D face landmarks) to $I$ while removing the influence of expressions. Our strategy is first to estimate the 3D positions of 2D landmarks with reconstructed face model parameters as shown in  Section~\ref{sec:overview}. Then a robust deformation transfer technique is proposed to determine the modified identity according to these 3D landmark positions as well as the expression. It is achieved by two steps: depth estimation for key points in 2D and solving an extended target function of~\cite{sumner2004deformation}. \emph{The main contribution of our identity deformation transfer algorithm is that we perform the deformation transfer between different feature spaces: from 2D to 3D; while prior work~\cite{sumner2004deformation, botsch2006deformation, baran2009semantic} performs the transfer in the same 3D space.}

To estimate the 3D position $(\hat{x}_i, \hat{y}_i, \hat{z}_i)$ of a certain modified landmark $\hat{\mathbf{l}}_i$ whose 2D coordinate in this frame is $(\hat{x}_i, \hat{y}_i)$, we need to reconstruct $\hat{z}_i$ (the depth of this point towards the screen plane). However, the depth value is unknown since the deformation is made in 2D space. Notice that when the front face is right against the screen plane, i.e., the face plane and screen plane are parallel to each other, points on the face mesh will have similar depth value (the delta of the maximum and minimum depth value of all the points will reach its minimum), especially for the landmark whose normal vector is perpendicular to the screen plane. Based on this observation, we estimate the depth $\hat{z}_i$ of the modified landmark $\hat{\mathbf{l}}_i$ with the original depth $z_i$ of $\mathbf{l}_i$, which can be computed directly according to Eq.~\ref{eq:fullfacemodel}. This estimation can avoid generating unrealistic facial identity effectively. However, as a result, our approach is able to achieve the best performance if the editing is applied on the frame when the actor is facing the camera.

Then we can get the modified facial identity by further transferring deformations (a set of modified 3D landmark coordinates computed above) to the target identity. Our approach is inspired by the correspondence system in~\cite{sumner2004deformation}, but developed in the context of our deformation framework. Let $V = \{\mathbf{v}_1, \dots, \mathbf{v}_n\}$ and $\hat{V} = \{\hat{\mathbf{v}}_1, \dots, \hat{\mathbf{v}}_n\}$ be the $n$ vertices of the original and modified facial identity. Note that here $V$ is equal to the facial identity $I$ when we remove the effect of the facial expression $E_t$ according to Eq.~\ref{eq:shapefacemodel}. We let $Q = \{\mathbf{Q}_i\}$ be a set of mesh triangles and $\mathbf{Q}_i = \hat{\mathbf{V}}_i(\mathbf{V}_i)^{-1}$ be the affine transformations that define the deformation for the $i$-th triangle, where $\hat{\mathbf{V}}_i$ and $\mathbf{V}_i$ are the corresponding vertex matrix~\cite{sumner2004deformation} calculated from $V$ and $\hat{V}$, respectively. The vertex positions of the modified identity are computed by minimize the distance between the original and modified 3D landmarks after removing the influence of facial expression. We define the landmark term as:
\begin{equation}
\label{eq:transfer:landmark}
\mathcal{E}_{l} = \frac{1}{2}\sum_{i = 1}^{m}||\hat{\mathbf{v}}_i - \tilde{\mathbf{l}}_i||^2,
\end{equation}
where $m$ is the number of modified 3D landmarks, and $\tilde{\mathbf{l}}_i = \hat{\mathbf{l}}_i + E_t^{(\mathbf{v}_i)}$ is the coordinate of the $i$-th modified landmark after merging the influence of the facial expression $E_t$. Then the whole energy function is defined as:
\begin{equation}
\label{eq:transfer:full}
\begin{array}{r@{\ }l}
\mathcal{E}(\hat{\mathbf{v}}_1, \dots, \hat{\mathbf{v}}_n) &= w_s\mathcal{E}_s + w_r\mathcal{E}_r + w_l\mathcal{E}_l \\[\jot]
\textrm{s.t.} \ \hat{\mathbf{v}}_i &= \mathbf{b}_k, \mathbf{b}_k \in \mathcal{F}(V),
\end{array}
\end{equation}
where $w_*$ controls the effect of each term, and $w_r = 0.1$ while the other two are set to $1$ experimentally; $\mathbf{b}_k$ in Eq.~\ref{eq:transfer:full} is a set of points on the face boundary $\mathcal{F}(V)$. $\mathcal{E}_s$ is a smooth term to make the transformations for the set of adjacent triangles $\mathcal{A}$ be similar with each other~\cite{sumner2004deformation}:
\begin{equation}
\label{eq:transfer:smooth}
\mathcal{E}_s(\hat{\mathbf{v}}_1, \dots, \hat{\mathbf{v}}_n) = \frac{1}{2}\sum_{i = 1}^{|Q|}\sum_{j \in \mathcal{A}(i)}||\mathbf{Q}_i - \mathbf{Q}_j||^2,
\end{equation}
and $\mathcal{E}_r$ maintains the original triangle shapes which are not affected by the deformation in order to prevent generating a drastic change in the shape of the target identity~\cite{sumner2004deformation}:
\begin{equation}
\label{eq:transfer:identity}
\mathcal{E}_r(\hat{\mathbf{v}}_1, \dots, \hat{\mathbf{v}}_n) = \frac{1}{2}\sum_{i = 1}^{|Q|}||\mathbf{Q}_i - \mathbf{I}||^2,
\end{equation}
where $\mathbf{I}$ is the identity matrix. The whole energy function can be minimized by solving a system of linear equations. During this procedure, the boundary points of the deformed mesh will match exactly since they are specified as constraints to keep the original face contour, the local deformations are transferred by the landmark term, and the rest of the mesh will be carried along by the smoothness and identity terms.

\section{Contour-based face model refinement}\label{sec:approach:refinement}

Since the quality of a 3D face model computed from the identity deformation transfer highly influences our final result, we implement an interface for artists to directly edit the deformed 3D face model for refinement. To ensure fluency and simplicity of user interaction during this process, we present a contour-based editing schema. Note that this refinement step can be skipped if a favorable result has already been achieved during previous steps or users have no experience in 3D model editing.

\begin{figure}[!t]
\centering
\includegraphics[width=\linewidth]{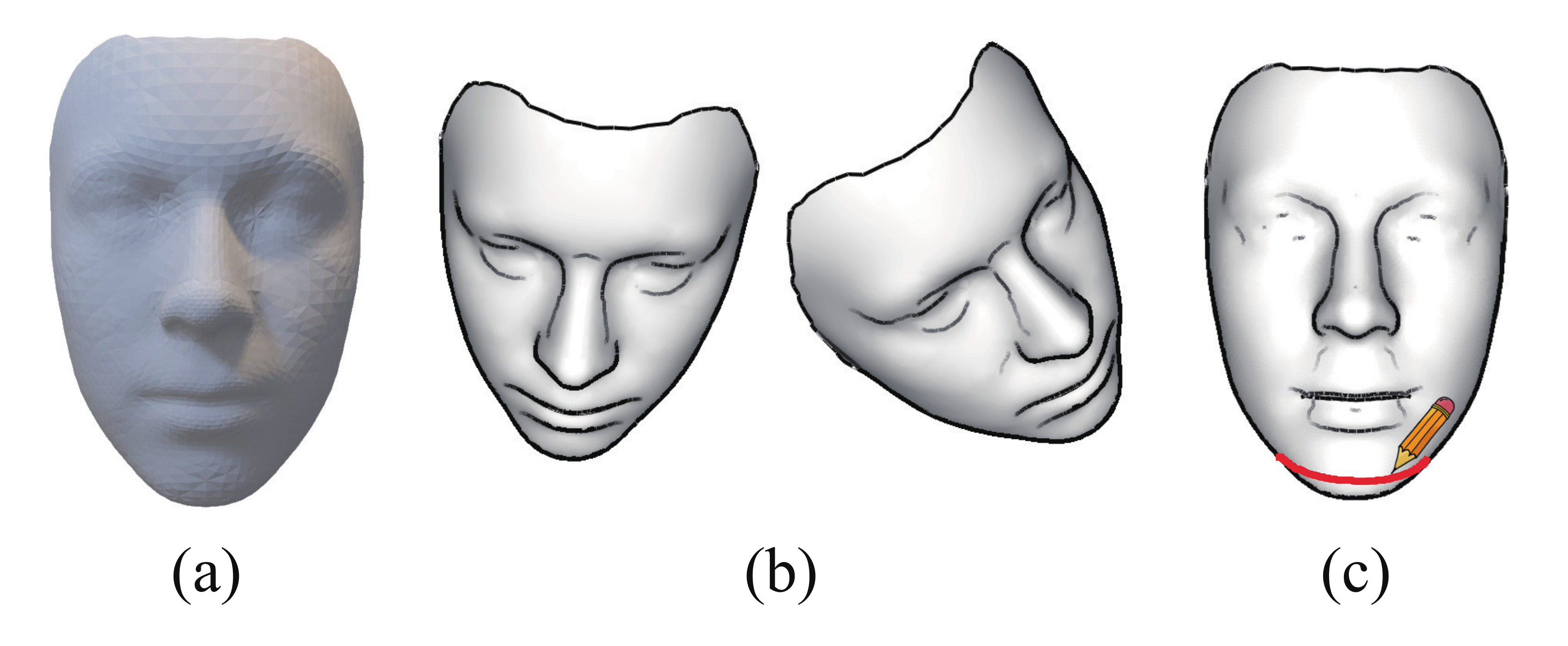}
\caption{We provide a contour-based mode for model refinement. Given a 3D face mesh (a), a hybrid line rendering method is utilized to render the contours from different view points (b). Users are allowed to edit these contours with sketches as shown in (c).}
\label{fig:refinement}
\end{figure}

At the beginning of this refinement stage, selected feature contours on the deformed 3D face model are projected onto the 2D canvas to produce an initial sketch-like contour map. Then the user can drag or rotate it to observe the model from different views. If some of the projected lines are not satisfactory, the user can redraw them with new sketches. In this refinement phase, these redrawn lines are matched more closely with user inputs by adding them as new constraints for the 3D face model. Compared with traditional interactive 3D mesh editing softwares such as MAYA or 3DS-MAX, which always take a long time for a skilled artist to create a decent 3D face model, our contour-based approach is much simpler and faster. Moreover, this contour-based refinement interface also ensures a smooth user experience during the whole editing process for not involving different softwares or interactive modes other than hand-drawn sketches.

\emph{Contour rendering.} Recent studies~\cite{wang2014anewsketch, zhao2015learning} present a hybrid line rendering method to generate 2D contour maps for 3D shapes and obtain good performance. In this paper, we adopt this approach which combines predefined exterior silhouettes, occluding contours, suggestive contours~\cite{decarlo2003suggestive} and shape boundaries to generate the final contour map from a given view point for further editing. Examples are shown in Fig.~\ref{fig:refinement}. Note that the preprocess steps in~\cite{zitnick2014edge} are also applied to reduce the noise in the initial map.

\emph{3D model refinement.} Users are allowed to rotate the input 3D model to edit its 2D contour map from different view points. Once an unsatisfactory line in the map is found, users can modify it by marking a certain region around to erase it first, and draw a new relevant sketch. After sampling the key points from them, the edited line is then calculated by the same algorithm as described in Section~\ref{sec:approach:sketch:fitting}. We treat the key points in the edited line as new landmark constraints, and the identity deformation transfer in Section~\ref{sec:approach:transfer} is employed to update the face model. Users are able to repeat these steps until a favorable 3D facial identity $\hat{I}$ is achieved. Note that since more constraints are added, we increase the weights of $w_s$ and $w_r$ in Eq.~\ref{eq:transfer:full} at this stage so as to prevent drastic deformations in the final 3D shape $\hat{I}$.

\section{Texture re-rendering and smoothing}\label{sec:approach:optimization}

Propagating deformations to the whole video is achieved by computing modified $\hat{F}_t$ with $\hat{I}$ for each frame $t$ according to Eq.~\ref{eq:fullfacemodel} and~\ref{eq:shapefacemodel}, and then re-rendering the extracted face texture isomap $M_t$ back to the frame with $\hat{F}_t$. For a high-fidelity result, the background should be warped as well, so that both sides of the face boundary deform coherently. We also apply the median filter on the boundary to blur the difference between the face and its surrounding background.

\begin{figure}[!t]
\centering
\includegraphics[width=\linewidth]{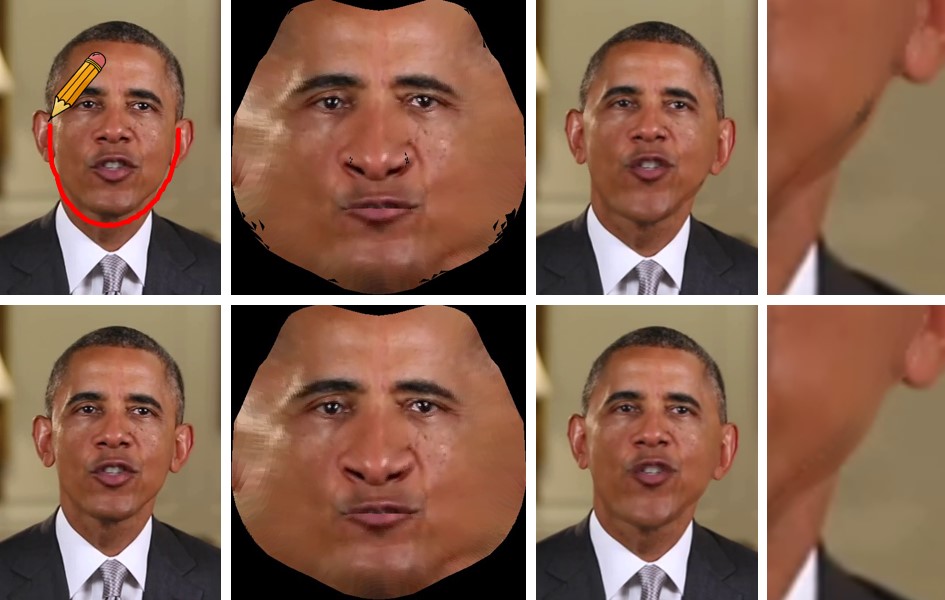}
\caption{We use the refined isomap (the second row) to remove artifacts. From left to right: the original frame and the sketch input, texture isomap, the modified output, detailed view of the output.}
\label{fig:isomap}
\end{figure}

Note that there might be some ``holes'' (invisible pixels) on the isomap $M_t$ due to the occlusion. However, if modifications are applied on the boundary of the 3D facial identity, these ``holes'' will lead to artifacts after smoothing since they might be visible as a result of the deformation. We notice that these missing pixels may be seen from other frames (since the actor always have different poses in different frames), which can be employed to fill ``holes'' in one certain frame. Therefore, we utilize a refined isomap $\hat{M}_t$ to synthesize the modified face in each frame. To obtain the refined isomap $\hat{M}_t$, we first compute a mean isomap $\bar{M}$ from all frames in the given video. And then for the isomap $M_t$ of each frame $t$, we use this mean isomap to fill the ``holes'' in it. We obtain the final refined isomap $\hat{M}_t$ for each frame after applying a Gaussian filter on it to smooth the boundaries. Finally, the artifacts on the boundary can be removed by re-rendering the face with $\hat{M}_t$ as shown in Fig.~\ref{fig:isomap}.

To handle missing background due to the facial deformation, one simple strategy is building a background model (background as in non-face and non-body) over successive frames and then replacing missing background pixels with newly revealed ones. However, this approach depends on an accurate background segmentation algorithm. In this paper, we solve it in a more robust warping-based manner. Firstly, we employ SIFT~\cite{lowe2004distinctive} to detect the static key points from the starting frame; optical flow is calculated throughout the whole video in order to track the dynamic key points of the background. Then we construct a set of control points for each frame by combining the static and dynamic points. Finally, we use Moving Least Square~\cite{schaefer2006image} algorithm to warp the background pixels based on detected control points. This optimization strategy can effectively avoid shaking for static objects in the background, while maintain the consistency of the face boundary concurrently.

\section{Results}\label{sec:results}

\begin{figure*}[!t]
\centering
\includegraphics[width=\linewidth]{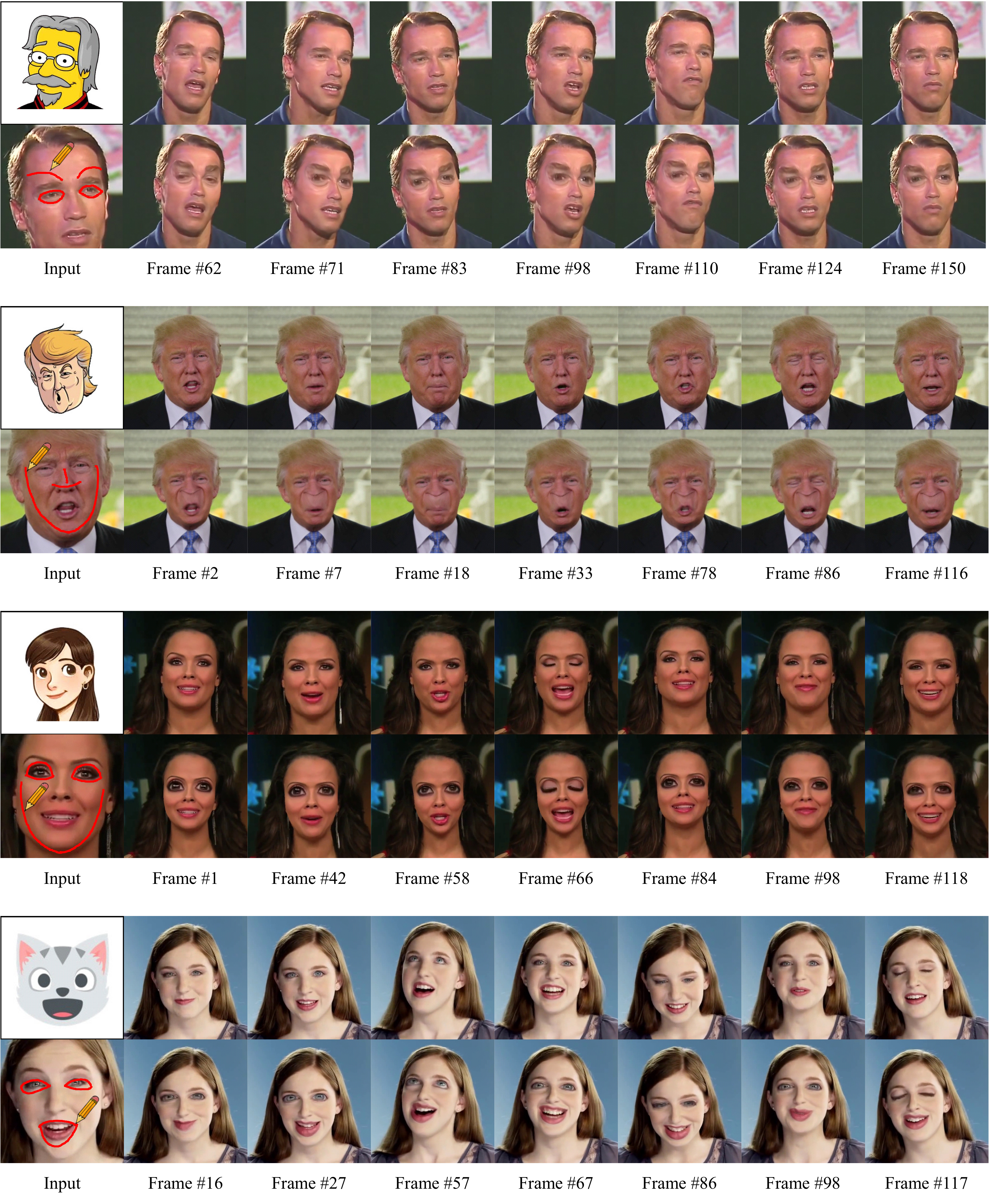}
\caption{More examples of our sketching system. In each example, the first column includes the cartoon image for reference (top) and the sketch input (bottom). Starting from the second column, the original (top) and edited frames (bottom) in different videos are shown.}
\label{fig:result}
\end{figure*}

We evaluate the performance of our approach on different Youtube videos at a resolution of $1280 \times 720$. The videos show different actors with different scenes captured from varying camera angles; we choose one frame for each video and provide a corresponding sketch of the actor's face as the input. In our experiments, users are allowed to edit 68 face landmarks marked by Huber et al.~\cite{huber2016multiresolution} by sketch for demonstration. Example results created by amateurs using our sketching interface are shown in Figs.~\ref{fig:teaser} and \ref{fig:result}.

\subsection{Runtime performance}\label{sec:results:runtime}

We evaluate the runtime performance of our methods by computing the average runtime of each step with respect to different video resolutions. Within the step of texture re-rendering, an isomap with $256 \times 256$ resolution is computed for 360P and 480P videos; $512 \times 512$ resolution is configured for HD videos. Our approach runs on a desktop computer with an Intel 4.00GHz Core i7-6700K CPU. Table~\ref{tbl:runtime} shows the result. The texture re-rendering and background optimization is the slowest components, while others run in a matter of milliseconds. Note that our framework can achieve real-time performance ($\geq 25$ FPS) for standard resolution videos without background optimization, which is compatible with streaming inputs. Moreover, our method does not rely on a power GPU and can be extended to light-weight devices.

\begin{table}[!t]
\renewcommand{\arraystretch}{1.2}
\caption{\label{tbl:runtime}Average runtime of our framework. The average runtime for one frame of each step in our method is computed with respect to different video resolutions. The program runs with parallel optimization.}
\centering
\begin{tabular}{@{}lcccc@{}}
\toprule
Video Quality & 360P & 480P & 720P & 1080P \\ \midrule
Model Fitting & 11.9ms & 12.1ms & 13.6ms & 14.3ms \\
Sketch Matching & 1.4ms & 1.4ms & 1.8ms & 2.1ms \\
Deform Transfer & 3.1ms & 3.1ms & 3.4ms & 3.5ms \\
Rendering & 16.6ms & 20.5ms & 82.6ms & 98.6ms \\
BG Opt. & 18.8ms & 21.6ms & 28.1ms & 34.2ms \\
FPS & \textbf{29.3Hz} & \textbf{26.9Hz} & 9.7Hz & 8.4Hz \\
FPS w/ BG Opt. & 19.2Hz & 16.8Hz & 7.6Hz & 6.2Hz \\
\bottomrule
\end{tabular}
\end{table}

\subsection{User study}\label{sec:results:userstudy}

In this section, the user study is conducted to evaluate the user experience as well as the video results achieved by our framework. We invited 20 people, 10 men and 10 women. All participants are graduate students aged from 23 to 25, and 3 of them major in arts while the others come from statistics and computer science departments. In addition, 4 of them have background in arts and 3D animation modeling (with 3-year experiences in MAYA and 3DS-MAX as well), while the others are amateur users that have limited or no knowledge about drawing and 3D editing. Before the following sessions, a 10-minute tutorial as well as 20 minutes for practice were given to guarantee that every participant knows how to use our interface. Another 40 minutes are employed to introduce MAYA to amateur users. In our evaluation, we use cartoon images as reference to compare our results with users' true editing intention in an effective way. Users are asked to edit the face to match a cartoon image using our system. However, they are free to make additional modifications as they wish. Therefore, the results may contain some unrelated user inputs.

\subsubsection{User experience of interfaces}

The goal of the first session is evaluating the user experience of our sketched-based interface. To guide users' editing intention, each participant was given a Youtube video together with a 2D cartoon face image as reference, and asked to edit the facial appearance of the actor in this video to match at least one prominent facial character in the cartoon image using our system. Note that the created facial identity was not required to strictly follow the reference image and differences were allowed. All participants should complete 5 tasks in this session with different pairs of videos and reference images as illustrated in Figs.~\ref{fig:teaser} and \ref{fig:result}. We also implemented another deformation-based user interface, where 3D face models were first calculated according to Section~\ref{sec:overview}; then MAYA was utilized as the editing tool instead of sketches, and final results were directly generated by the modified models as Section~\ref{sec:approach:optimization}. In this deformation-based interface, users are only allowed to edit/modify the positions of mesh vertices with MAYA. This constraint is made for fair comparison, since our system deforms the mesh in the same way. All participants were asked to repeat the same tasks with this interface. For amateur users, they could stop anytime if it was too difficult for them to continue while the artists were required to finish all tasks. To verify the effectiveness of our contour-based refinement mode in Section~\ref{sec:approach:refinement}, the amateur users were recommended to try it after the initial sketching, and the artists should use it for all tasks. At the end of this session, we asked the participants which system is easier to use.

Among all the 16 amateur users, 12 of them finished editing tasks with the deformation-based interface while the others gave up halfway. Instead, all participants completed the same tasks with our sketch-based interface. Moreover, 10 of the amateurs used our contour-based refinement mode; the others chose to skip it for favorable results had already been achieved in the previous steps. In terms of user experience, all participants agree that our system is simpler to use and yields decent results. Those who tried our refinement mode agree that it was very helpful, and the four professional artists agree that it is more efficient than traditional deformation-based software.

\subsubsection{Comparison on mesh deformation}

\begin{figure}[!t]
\centering
\includegraphics[width=\linewidth]{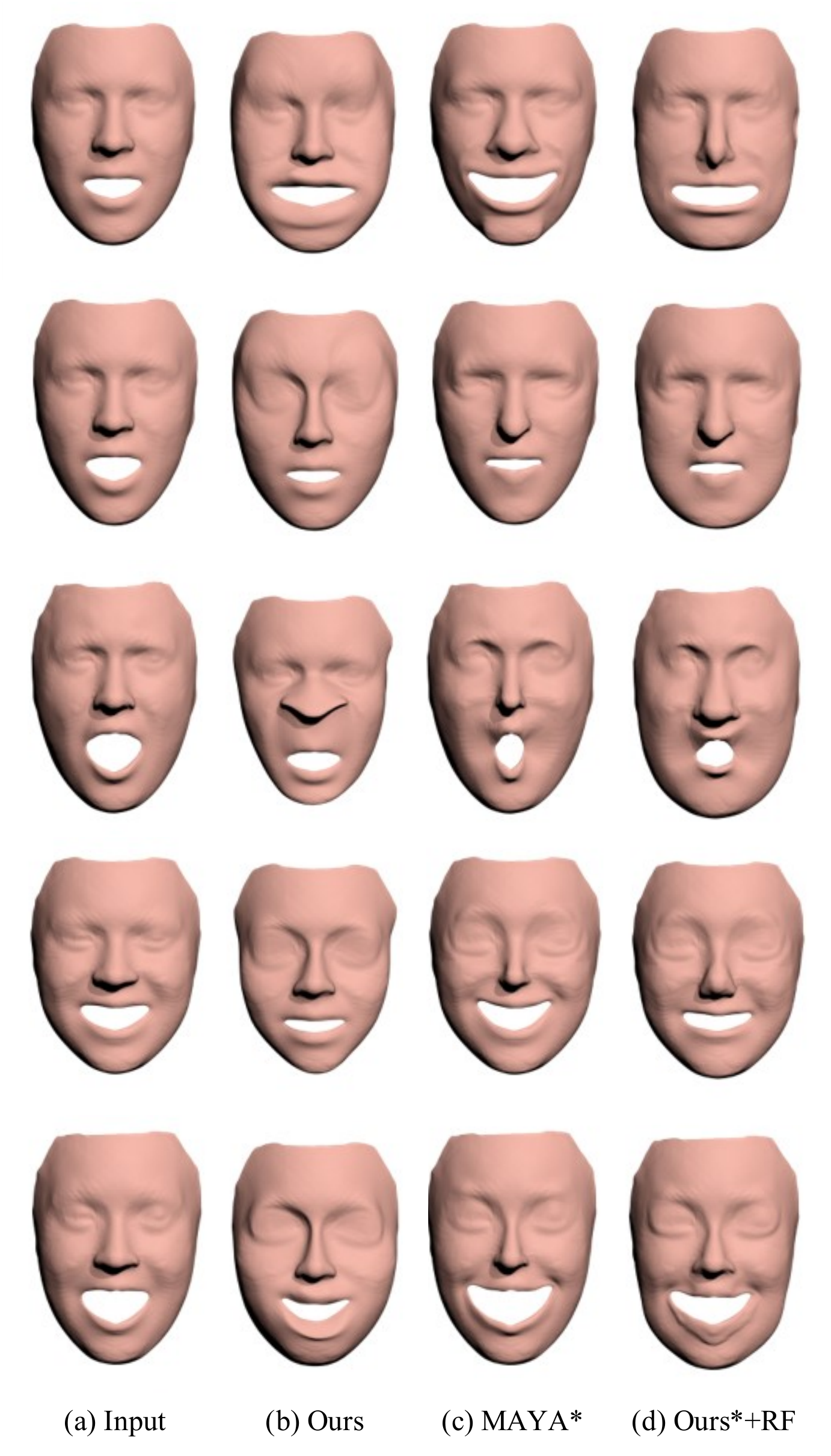}
\caption{A gallery of results created using three different editing interface settings. From top to bottom, each row shows the fitted 3D face model and the deformed models corresponding to the cases as shown in Fig.~\ref{fig:teaser} and Fig.~\ref{fig:result}, respectively. Note that each 3D face model is fitted from the actor in the first frame of the video.}
\label{fig:deformation}
\end{figure}

Fig.~\ref{fig:deformation} shows a gallery of deformed 3D face models, which are corresponding to the actors of the cases as shown in Figs.~\ref{fig:teaser} and \ref{fig:result}, with three different editing interface settings. They were created using our sketch-based interface by amateurs~(Ours), by artists with the deformation-based interface~(MAYA*) and our sketch-based interface with the refinement mode~(Ours*+RF), respectively. Detailed timings are shown in Table~\ref{tbl:usertime}. In addition, we also report the timing for editing using the deformation-based interface by amateurs~(MAYA). As shown in Table~\ref{tbl:usertime}, an amateur on average only spent 3.6 minutes to complete the task via our system, which is 3 times faster than using the interface based on MAYA. Meanwhile, our sketching interface also doubled the editing efficiency for artists. In Fig.~\ref{fig:deformation}, our system is able to achieve results comparable with the ones created using MAYA by artists, while amateurs managed to perform reasonable mesh deformations with our interface.

\begin{table}[t]
\renewcommand{\arraystretch}{1.2}
\caption{\label{tbl:usertime}Average timings for editing with different interfaces.}
\centering
\begin{tabular}{@{}lllll@{}}
\toprule
Interface & MAYA & Ours & MAYA* & Ours*+RF \\ \midrule
Time (m) & $12.5 \pm 1.8$ & $3.6 \pm 0.3$ & $9.5 \pm 1.4$ & $4.1 \pm 0.8$ \\
\bottomrule
\end{tabular}
\end{table}

\subsubsection{Evaluation on visual results}

\begin{figure}[!t]
\centering
\includegraphics[width=\linewidth]{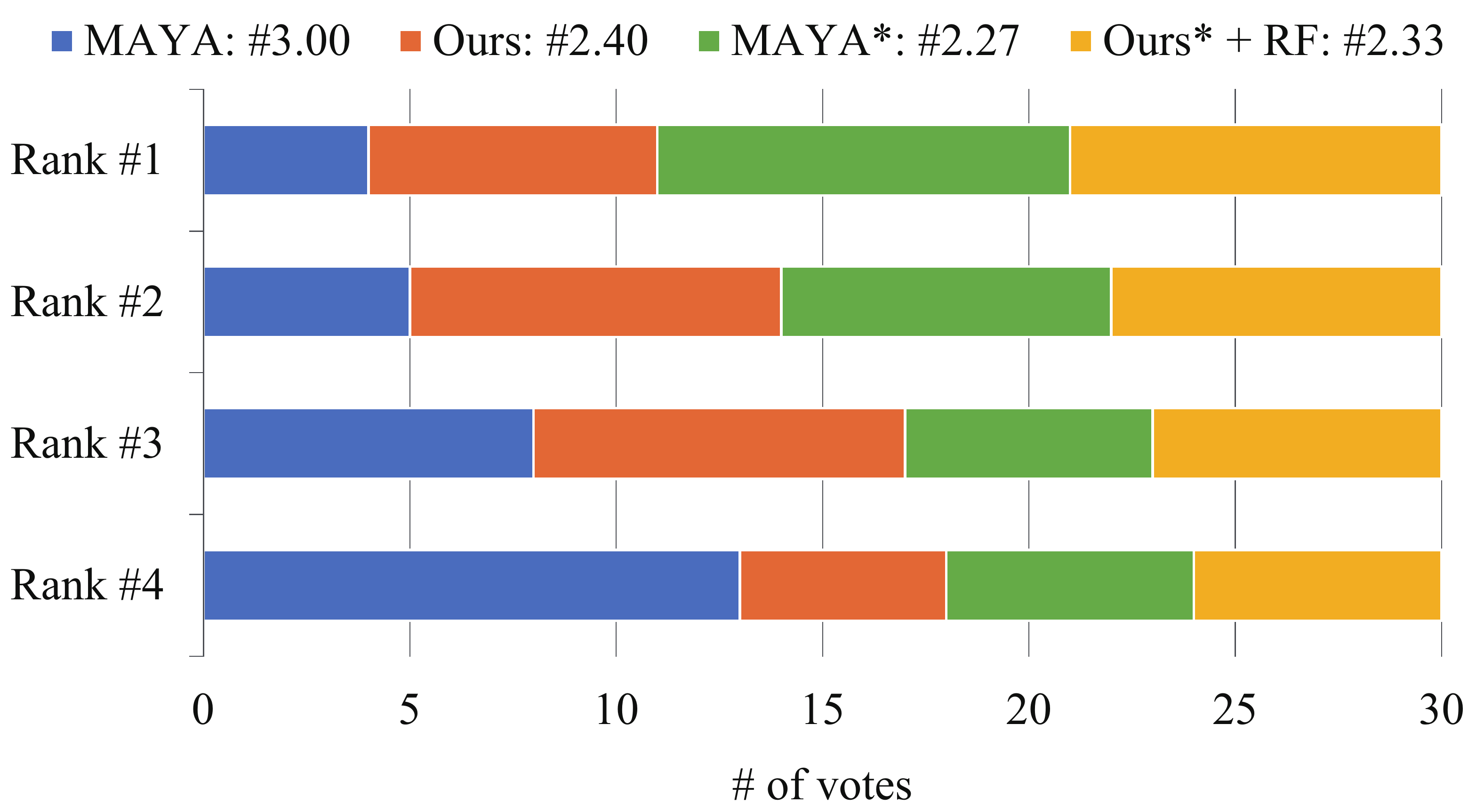}
\caption{Voting results from the 1st to 4th rank of videos created with 4 different interface settings. We also report the average ranking of each group following the group name.}
\label{fig:userstudy}
\end{figure}

To further evaluate the visual results generated by our sketch-based face editing system, we also designed the second session following the editing session. In this session, results in the previous session were selected into 4 groups: videos edited by amateurs using the deformation-based interface or our sketch-based interface, and ones generated by artists using the deformation-based interface or our sketch-based interface with the refinement mode. For fair comparison, we manually chose the best result in each group for every task in the editing session. After that, all these videos (5 tasks done by 4 groups, respectively) together with their corresponding reference cartoon images were presented to additional 30 students who did not participated in the editing session. Given a reference image (displayed in random order), every participant was asked to look at the corresponding videos from the 4 different groups, and then rank them by choosing the one that better matches the cartoon. The final results are shown in Fig.~\ref{fig:userstudy}.

We can find that videos created by amateurs using our sketching interface obtain higher average rank than ones using deformation-based interface. A T-Test was also conducted to compare rankings obtained with these two different settings. There was a significant difference in the rankings using the deformation-based interface ($M = 3.00$, $SD = 1.17$) and sketch-based interface ($M = 2.40$, $SD = 1.08$); $t(58) = 2.19$, $p = 0.03$. It demonstrates that our framework does improve the results for amateurs. Notice that different results created by artists and amateurs with our sketching interface have similar average ranks (2.33 and 2.40, respectively). It suggests that, with the help of our sketch-based interface, amateurs manage to create comparable results to a certain extent compared with artists. Another observation is that results have similar ranks when artists use our interface and MAYA, which agrees that artists can produce competitive results with our interface compared with MAYA.

\subsection{Evaluation of 3D-based editing}

We argue that the 3D face model and deformation transfer algorithms are the keys to ensure the consistency as well as fidelity of the editing results. To evaluate the effectiveness of our 3D-based editing system, we implemented a 2D-warping baseline for comparison. We use the results of deformed 2D face landmarks in Section~\ref{sec:approach:sketch:fitting} as the inputs for this baseline. Then the face is deformed by Moving Least Square~\cite{schaefer2006image} in the 2D space. All participants of the user study are invited to try this 2D-warping baseline, and 85\% of them prefer our system over the baseline.

For quantitative comparison, we measure the content consistency of an editing video using the Content Distance metric introduced in~\cite{tulyakov2017mocogan,zhao2018learning}. We employed OpenFace~\cite{amos2016openface}, which outperforms human performance in the face recognition task, for measuring video content consistency. A feature vector is produced by OpenFace for each frame in a given video. The distance is then calculated by the pairwise L2 distance of the feature vectors. To measure the quality of an edited video, we compute the distances between each frame and the first frame. A method that owns a lower distance curve handles changes in rotations or expressions better.

\begin{figure}[!t]
\centering
\includegraphics[width=\linewidth]{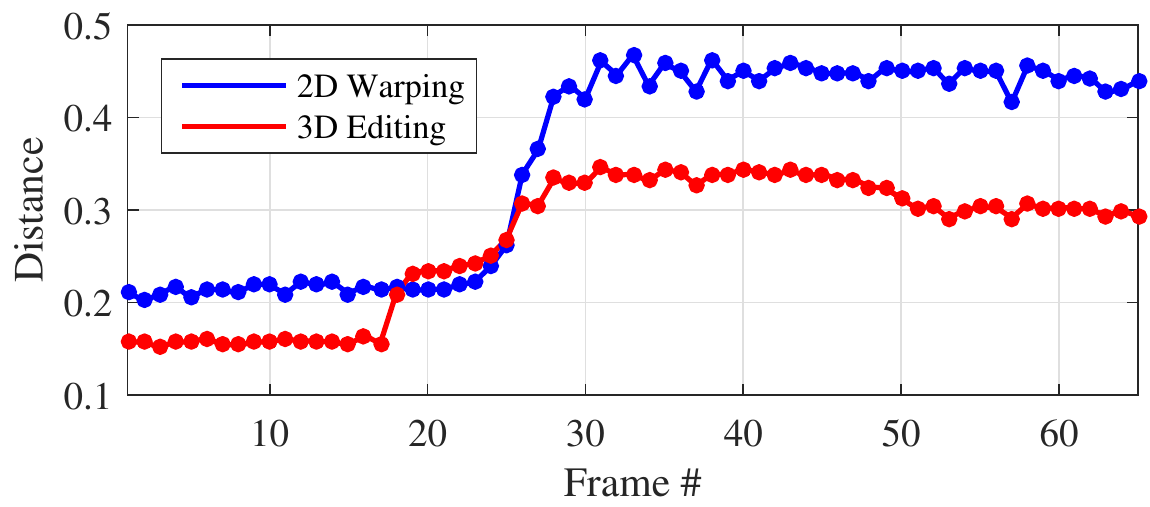}
\caption{The Content Distance curves of our 3D-based editing system and the 2D-warping baseline. Our system has a lower curve and a smaller gap when handling changes in head rotations and facial expressions.}
\label{fig:3dediting}
\end{figure}

We collected all the edited results in the user study for evaluation. To highlight the performance, the videos are manually aligned so that a noticeable change in head rotation or expression occurs around the 25th frame. Using the deformed 2D landmarks as the inputs, we compute the distance curves for the edited videos generated by our 3D-editing system and the 2D-warping baseline for comparison. The results are given in Fig.~\ref{fig:3dediting}. From the figure, we find that the content of the videos generated by our system is more consistent: ours achieves a lower distance curve. More importantly, our system is able to handle rotation or expression changes better than the 2D-warping baseline, since there is a smaller gap in our distance curve. Therefore, our system offers a 3D solution which substantially outperforms the 2D-warping approach.

\subsection{Evaluation of sketch matching}

\begin{figure}[!t]
\centering
\includegraphics[width=\linewidth]{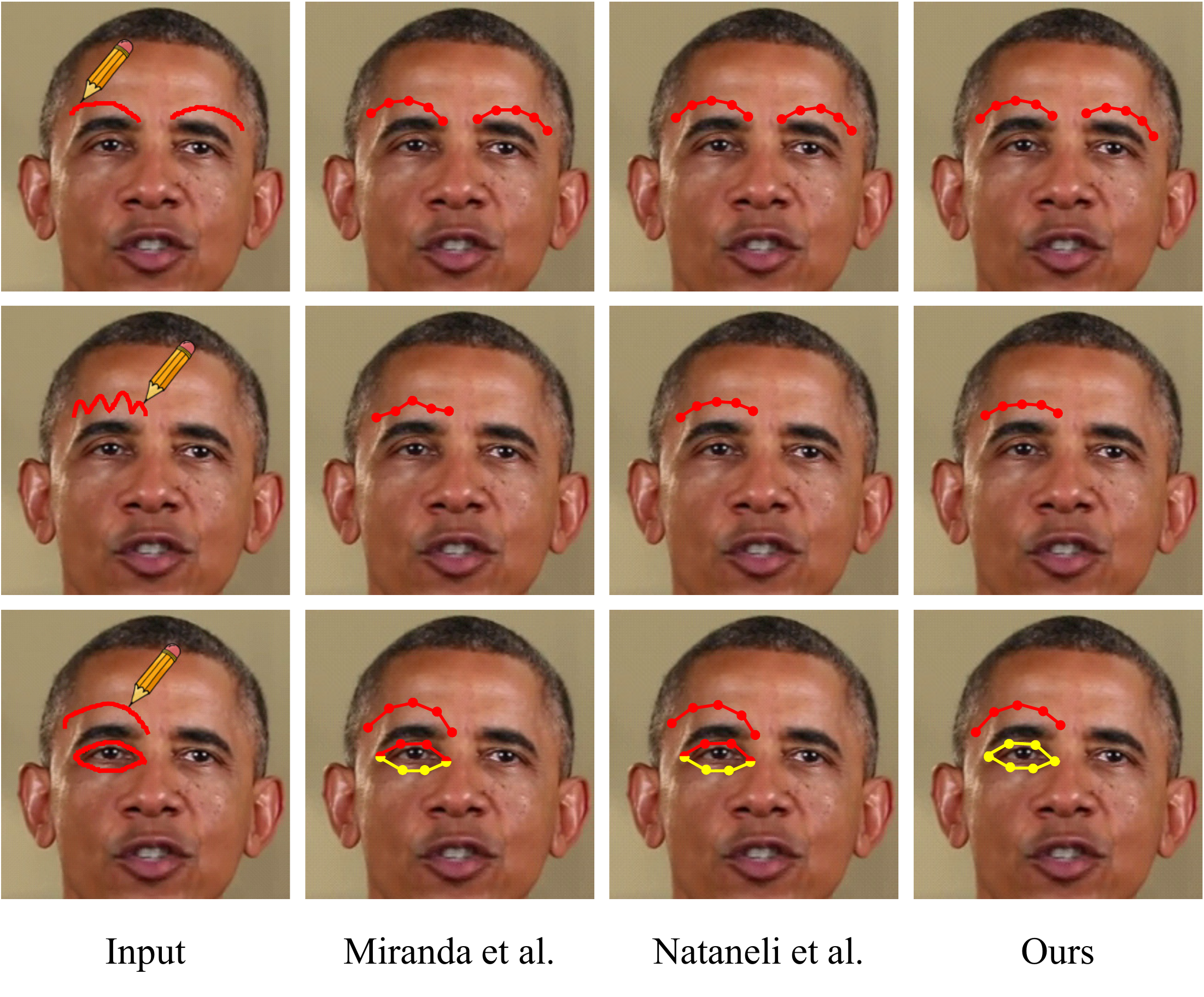}
\caption{Comparison of our sketch matching to the approximate implementation of Miranda et al.~\cite{miranda2012sketch} and Nataneli et al.~\cite{nataneli2011bringing}. From top to bottom: two individual strokes, one stroke with noises, and a group of strokes with ambiguity (different colors represent different matched landmark group pairs). Our method is able to achieve reasonable results for all these cases.}
\label{fig:results:sketch}
\end{figure}

To evaluate the performance of sketch matching, we compare our method with a geometry-based algorithm described in~\cite{miranda2012sketch} and another learning-based approach~\cite{nataneli2011bringing} which both achieve state-of-the-art performance. We use the stroke similarity measurement described in them to match strokes with landmarks, respectively as their corresponding approximate implementation. Detailed results are shown in Fig.~\ref{fig:results:sketch}. We can find that all the methods produce competitive results with clear user inputs. However, as shown in the second case of Fig.~\ref{fig:results:sketch}, \cite{miranda2012sketch} is sensitive to noises since it fits landmarks to strokes via only geometry features; Nataneli et al.~\cite{nataneli2011bringing} is able to handle this case due to pre-learned prior knowledge; our method can remove noises by taking the original appearance of the landmark group (the shape prior term in Eq.~\ref{eq:sketch:fitting:target}) into consideration. For ambiguous inputs as shown in the third case of Fig.~\ref{fig:results:sketch}, both~\cite{miranda2012sketch} and~\cite{nataneli2011bringing} map the second stroke to eyebrow incorrectly; we can successfully match it with the upper eyelid since the HMM we employed trends to match the upper and bottom eyelids at the same time during optimization.

\section{Conclusions and discussion}\label{sec:conclusion}

This paper presents the first sketch-based face editing framework for monocular videos. In an attempt to recognize the user's editing intentions from hand-drawn sketch, a robust sketch matching schema is introduced to convert them to a set of face landmark deformations. Furthermore, a novel facial identity deformation transfer algorithm is employed to propagate these changes throughout the whole video, while consistency and fidelity are maintained. Without background optimization, our framework is able to achieve real-time performance for streaming inputs with standard definition. Overall, we believe our framework will contribute to many new and exciting applications in the field of face editing on light-weight devices, e.g., a tablet PC and mobile phone.

\emph{Limitations.} There are some notable limitations to our work. One limitation of our sketch-mapping algorithm is that amateur users may not produce correctly ordered sketches in their first try. We make use of HMM to model the relation of different strokes, and users have to redraw a part of strokes with the incorrect order. However, this is mitigated by the fact that on average, the complexity and number of sketches is small (less than 6 strokes in most cases), and our interactive system supports rapid iteration and refinement of strokes. Another limitation is that we only allow users to edit a few landmarks on a face. This is due to the limitation of the morphable models we employed to construct the 3D face: local geometric details such as wrinkles cannot be represented. Moreover, since our method relies on a fixed z-value for more accurate depth estimation, users have to draw sketches on a front face to achieve the best performance.

\emph{Future work.} In the future, we will consider implementing a stereo editing interface to enhance the user experience and enable users to edit faces from different view points as well. To alleviate the problem of limited editing ability, we propose to utilize Generative Adversarial Network~(GAN)~\cite{ian2014generative} to make pixel-to-pixel prediction~\cite{zhao2018learning,isola2017image,tian2018cr} directly from sketches instead of using morphable models. Moreover, allowing users to edit more facial details from sketches is another future direction. We expect other interesting applications for the framework we have shown here. One can imagine coupling this work with an artist to create a cartoon talking avatar starting from a video.

\section*{Acknowledgments}
The authors would like to thank the reviewers for their constructive comments and the participants of our user study for their precious time. This work was funded in part by grant BAAAFOSR-2013-0001 to Dimitris Metaxas. This work is also partly supported by NSF 1763523, 1747778, 1733843 and 1703883 Awards. Mubbasir Kapadia has been funded in part by NSF IIS-1703883, NSF S\&AS-1723869, and DARPA SocialSim-W911NF-17-C-0098.

\bibliographystyle{cag-num-names}
\bibliography{refs}

\end{document}